\documentclass[11pt,a4paper]{article}
\usepackage{xr-hyper}
\usepackage[hyperref]{emnlp2020}
\usepackage{times}
\usepackage{latexsym}
\usepackage{placeins}

\usepackage{longtable}

\usepackage{microtype}
\usepackage{wrapfig}
\usepackage{tabularx}
\newcolumntype{Y}{>{\centering\arraybackslash}X}
\usepackage{booktabs}
\usepackage{graphicx}
\usepackage{verbatim} 
\usepackage[utf8]{inputenc}
\usepackage{booktabs}
\usepackage{multirow}

\aclfinalcopy 
\def\aclpaperid{2071} 

\newcommand{\stgur}{Essays}
\newcommand{\asrd}{ASRD}
\newcommand{\holj}{HOLJ}
\newcommand{\tos}{ToS}
\newcommand{\grasp}{GrASP}
\newcommand{\geb}{GE}
\newcommand{\beg}{Split}
\newcommand{\graspl}{\grasp$^{\hspace{1pt}\scriptsize{\textrm{\textit{lite}}}}$} 
\newcommand{\as}[1]{[\texttt{#1}]}
\newcommand{\sr}[3]{$#1\!#2\!#3$}   

\title{Unsupervised Expressive Rules Provide Explainability and \\ Assist Human Experts Grasping New Domains}

\author{Eyal Shnarch\thanks{\ \ First two authors equally contributed to this work.}, Leshem Choshen\footnotemark[1], Guy Moshkowich, Noam Slonim, Ranit Aharonov
\\
IBM Research \\
\\
\{eyals, leshem.choshen, noams\}@il.ibm.com,
\\
\{guy.moshkowich, ranit.aharonov2\}@ibm.com  }
  
\date{}

\begin{document}
\maketitle
\begin{abstract}

Approaching new data can be quite deterrent; you do not know how your categories of interest are realized in it, commonly, there is no labeled data at hand, and the performance of domain adaptation methods is unsatisfactory.  

Aiming to assist domain experts in their first steps into a new task over a new corpus, we present an unsupervised approach to reveal complex rules which cluster the unexplored corpus by its prominent categories (or facets). 

These rules are human-readable, thus providing an important ingredient which has become in short supply lately - explainability. Each rule provides an explanation for the commonality of all the texts it clusters together.

We present an extensive evaluation of the usefulness of these rules in identifying target categories, as well as a user study which assesses their interpretability.

\end{abstract}

\vspace{-0.2cm}
\section{Introduction} \label{sec:intro}

A common scenario faced by subject matter experts tackling a new text understanding task is getting to know a new dataset, for which there is no labeled data. Understanding the unexplored data, and collecting first insights from it, is always a slow process. Often, the expert is trying to categorize the data, and potentially build a system to automatically identify these categories. For example, an expert may be looking at customer complaints, aiming to understand their types or categories, and then building a system that will categorize complaints. Or she may be analyzing contracts, aiming to identify the types of legal commitments. 

In other cases, the expert may be trying to identify a certain class of texts, and this class may be composed of unknown underlying sub-types or categories. Consider a data scientist looking for all arguments, related to a suggested policy, raised in a public participation forum. These arguments may be of several types, which are a-priori unknown.

When facing a new task, with no labeled data, but with domain expertise, a practical first step is to manually write rules that identify some texts from a certain category the expert is aware of and aiming to identify
(e.g., a certain complaint type). With these seed examples, experts can better understand the occurrences of the target category in the new corpus, and use them as the 
initial set 
of labeled examples, towards the goal of having enough labeled data to 
facilitate 
supervised learning.

However, oftentimes, the categories underlying the data are not known a-priori, and may be a part of what the expert aims to identify (e.g., what are the types of complaints). Since new data may mean new underlying categories, domain adaptation is not always applicable, and often results in unsatisfying performance \cite{ziser2018pivot}.

\begin{table*}[t]
 \resizebox{\linewidth}{!}{
\begin{tabular}{@{}cclc@{}}
    Dataset-category & No. &  Sentences Matched & Rule  \\ 
    \cmidrule[\heavyrulewidth](l){1-4}    

    \begin{tabular}{@{}c@{}} \asrd{}\\ argument\end{tabular} &
    \begin{tabular}{@{}c@{}} 1 \\2\\3 \end{tabular} &
     
    \begin{tabular}{@{}l@{}} so \textbf{first} let us \textbf{address} the question \\ 
    our \textbf{second argument} is about \\ 
    my \textbf{first overview} is \end{tabular} &
     
     \begin{tabular}{@{}c@{}} \as{hyponym of rank} $+$\\ \as{WordNet super class of communication} \\
     \\
     (an ordinal number,
      a term relating to human communication)
     \end{tabular}  \\ 
    \cmidrule[\lightrulewidth](l){1-4}
    
    \begin{tabular}{@{}c@{}} \asrd{}\\ argument \end{tabular} &   
    \begin{tabular}{@{}c@{}} 4 \\5\\6 \end{tabular} &

    \begin{tabular}{@{}l@{}} \textbf{additionally I think} that sam is \textbf{confused}    \\
    \textbf{ultimately}, \textbf{we think} that it \textbf{limits} the \\
    \textbf{obviously}, \textbf{we acknowledge} it's \textbf{important} \end{tabular} &
    
    \begin{tabular}{@{}c@{}} \as{adverb} $+$ \as{personal pronoun} $+$ \\ \as{hyponym of think} $+$ \as{sentiment word}\\
    \\
    (an adverb, 
    an indication of a person, \\
    a term related to thinking,
    and a word bearing a clear sentiment)
    \end{tabular} \\
   
    \cmidrule[\lightrulewidth](l){1-4}
    
    \begin{tabular}{@{}c@{}} \stgur{}\\ premise \end{tabular} &   
    \begin{tabular}{@{}c@{}} 7 \\8\\9 \end{tabular} &
    
    \begin{tabular}{@{}l@{}}     
    \textbf{for example}, \textbf{employer} always prefer to\\
    \textbf{for instance}, several \textbf{teenagers} play games\\
     as a matter \textbf{of fact}, \textbf{women} have proved
     \end{tabular} &
    
    \begin{tabular}{@{}c@{}} \as{preposition} + \\ \as{hyponym of psychological feature} + \\ \as{hyponym of causal agent} \\
    \\
    (a preposition, a term related to the mental domain, \\
    and an entity that can cause a change of any type)
    
    \end{tabular} \\
    
    \cmidrule[\lightrulewidth](l){1-4}
      
    \begin{tabular}{@{}c@{}} \holj{}\\ background \end{tabular} & 
    \begin{tabular}{@{}c@{}} 10 \\11\\12 \end{tabular} &
    
    \begin{tabular}{@{}l@{}}    
    \textbf{Section} 171B ( 1 ) \textbf{provides} :
     \\ " ( 1 ) This \textbf{section} \textbf{applies} where -
     \\ \textbf{Paragraph} 11 of the circular \textbf{states} :
      
     \end{tabular} &
    
    \begin{tabular}{@{}c@{}} \as{hyponym of written communication} + \\ \as{noun} + \\ \as{Verb, 3rd person singular present} \\
    \\
    (a written entity, followed by a noun, \\
    and a verb for he/she/it in present tense)

    \end{tabular} \\
    
    \cmidrule[\lightrulewidth](l){1-4}
      
    \begin{tabular}{@{}c@{}} \tos{}\\ unfair clauses  \end{tabular} & 
    \begin{tabular}{@{}c@{}} 13\\14\\15 \end{tabular} &
    
    \begin{tabular}{@{}l@{}}
    we may take any of \textbf{these} actions at \textbf{any} time
     \\ suspend \textbf{your} access to \textbf{any} of the
     \\ \textbf{no} liability to you or \textbf{any} third party
      
     \end{tabular} &
    
    \begin{tabular}{@{}c@{}} \as{ndet syntactic relation}  +  \as{any} \\
    \\
    (a noun determiner, followed by the word ``any'')
    
    \end{tabular} \\
    
    \cmidrule[\heavyrulewidth](l){1-4}
\end{tabular}
}

\caption{Examples for rules and generalizations found. Matched words are in bold. A description of each rule is provided below it, in parentheses. The datasets and categories are described in Section \ref{sec:data}.}
\label{tab:pattern_exmples}
\vspace{-0.4cm}
\end{table*}

In this paper, we present a method for generating initial rules automatically, with no need for any labeled data, nor for a list of categories. 

Our method, \graspl{}, is based on \grasp{} \cite{grasp}. \grasp{} is a supervised algorithm that finds highly expressive rules, in the form of patterns, that capture the common structures of a category of interest. \grasp{} requires a set of texts in which the target category appears and a set in which it does not. \graspl{} is an unsupervised version of \grasp{}, that requires no labeled data and no prior knowledge. 

Instead, \graspl{} takes a \textit{background} corpus and contrasts it with the new corpus, the \textit{foreground} corpus. By this, it reveals rules which capture sentences that are common in the foreground but not in the background. Such sentences are expected to be correlated with (at least some of) the unique categories in the foreground -- the new corpus. Examples of such rules are given in Table \ref{tab:pattern_exmples}.

Naturally, rules generated without supervision would be noisy. In addition, the rules revealed by \graspl{} capture a mixture of the categories that exist in the foreground corpus, some of which may be irrelevant for the task at hand. We, therefore, suggest \graspl{} as a preliminary automatic step which provides input for the human expert, without any input needed beyond the corpus of text. As rules are human-readable, and each one provides an explanation for why it clusters sentences together, experts can identify the subset of rules which, together, best capture the sentences of their category of interest. Experts can also  be inspired by the rules suggested by \graspl{}, manually edit rules to better fit their needs, merge elements from several rules into new rules, or improve their own manual rules with generalizations offered by the suggested rules. In other words, \graspl{} is a way to alleviate the 
blank canvas problem 
and to expedite the expert's work.

The rules identified by \graspl{} not only elucidate the underlying categories and facilitate rule-based algorithms, but also provide the benefit of explainability. That is, 
the human expert can now explain why a text is classified as a complaint and why it is in a certain complaint category.

We extensively evaluate \graspl{} over 
datasets from different domains, and show that the rules it generates, without being exposed to the datasets' categories, can help identify  
these categories. We further present a user study which validates the explainability power of \graspl{} rules.

\vspace{-0.1cm}
\section{\grasp$^{\hspace{1pt}\small{\textrm{\textit{lite}}}}$} \label{sec:method}

When facing a new task with new data, it is useful to have a tool which can quickly highlight some interesting aspects of these data.  
Such a tool must work with minimal prerequisites, as often we have little information about the new data. 

This is what our proposed method, \graspl{}, aims to provide. \graspl{} is based on \grasp{} \cite{grasp}, an algorithm for extracting highly expressive rules, in the form of patterns, for detecting a target category in texts.

A good rule is one that captures different realizations of the target category. For example, humans reading 1--3 in Table \ref{tab:pattern_exmples} can notice their commonality, even if they cannot name it. \grasp{} offers a rule which generalizes these realizations, and reveals their common structure: a hyponym of the noun \emph{rank}, closely followed by a noun which is a descendant, in WordNet, of \emph{communication}.   

To achieve this goal, \grasp{} extracts patterns that characterize a target linguistic phenomenon (e.g., argumentative sentences). Its input is a set of positive examples (in which the phenomenon appears) and another set of negative examples (in which it does not). First, all terms of all examples are augmented with a variety of linguistic attributes. Attributes are any type of term-level information, such as syntactic information (e.g., part-of-speech tag, information from the parsed tree), semantic (e.g., is the term a named entity? what are its hypernyms?), task-specific (e.g., is the term included in a relevant lexicon?), and more.
Next, GrASP greedily selects top attributes according to their information gain with the label. These attributes make the alphabet. Patterns are grown in iterations by combining attributes of the alphabet with shorter patterns from the previous iteration. At the end of each iteration a greedy step keeps only the top patterns (by information gain). 

In this work, we use a commonly available attribute set, which includes the surface form of the term, POS tags, Named Entity Recognizer, 
WordNet \cite{miller1998wordnet}, and a sentiment lexicon. We used the same set of attributes throughout our experiments, but one can add specific ones or rely on different technologies to extract them (e.g., a new parsing technology). See rule examples in Table~\ref{tab:pattern_exmples}. 

As the rules are human-readable and expose common structures in the data, they can expedite the process of getting to know it, especially when addressing novel domains. 

An entry barrier is that \grasp{} requires
labeled data which may not be available for a new domain. \graspl{} aims to lift this barrier
by providing a method to generate the two input sets for GrASP, with no labeled data.
It achieves that by setting a more modest goal -- instead of discovering rules describing common structures of a target category, \graspl{} aims to discover rules describing non-trivial structures which capture some repeating meaning, or category. However, these rules must not overfit the available data.

To achieve this goal, \graspl{} contrasts the available data, the \textit{foreground corpus} (which serves as the positive set), with a \textit{background corpus} (used as the negative set) in which the categories of interest are expected to be significantly less prominent.
With these two input sets, the regular \grasp{} can be applied.
By the nature of weak supervision, the foreground is not guaranteed to contain only positive examples (same for background and negative). However, we hypothesize that it is enough for a phenomenon to be more prominent in the foreground than it is in the background, for the regular \grasp{} to extract rules that characterize it. 
This way, by discovering rules for repeating meaningful structures which tend to appear in the foreground corpus more than in the background corpus, \graspl{} describes the common and unique categories of the available data. Next, we describe two methods to obtain a background corpus.

\paragraph{General English} A simple choice is to take random texts of the language of interest. We sampled 50,000 sentences from a news-domain corpus. In many cases, such a corpus is, on the one hand, different enough from the domain corpus (so can be assumed to be less enriched with the target category), and on the other hand, similar enough so as not to make the discrimination task of \graspl{} trivial (which will result in non-informative rules). 

However, in other cases, such a random sample of texts would not yield a suitable background corpus. For a distinctive domain corpus, legal contracts for example, contrasting it with a general English background will mostly bring up the legal jargon which is very common in the domain and rare in general English. The structures of legal commitments, a potential
target category, would be obscured by this specificity of the domain. Thus, another method is needed, one which builds a background corpus from the domain corpus itself.

\paragraph{In-Domain Split} For those cases, in which a general English background is too distinct from the foreground corpus, we suggest splitting the domain corpus itself into foreground and background. 
In this in-domain split the language style in the foreground and the background are similar, thus it avoids the risk of discovering rules that simply capture stylistic differences between the two parts.

If the expert has some knowledge about the new domain, it can be used to come up with a heuristic to split the new corpus. As an example of knowledgeable in-domain split we take the argument mining task. Argumentative sentences, aiming to persuade, ought to be well structured, to be easily understood by an audience, and often include foreshadowing hints, to guide the audience through the full argument. We hypothesize that such structures are more likely to be found in the beginning of a sentence, rather than in its end.
Based on this hypothesis, the foreground is made of the first halves of all sentences in the corpus, while the background is made of the second halves. We used this split method as an example in the analysis in \S\ref{ssec:split_by_indicator}.

If no heuristic can be found for the dataset, we suggest splitting it based on an unsupervised clustering method. The expert examines the clusters and chooses as the foreground one cluster which seems to contain many sentences of the target category. 
This selection does not have to be optimal (i.e., choosing the cluster with the most relevant sentences). It is enough that the prior for the target category in the selected cluster would be considerably higher than the prior in the entire corpus. The rest of the clusters are used as the background.

\vspace{-0.1cm}
\section{Datasets} \label{sec:data}
\vspace{-0.1cm}

To demonstrate that \graspl{} rules are useful across domains, we evaluate them on 10 datasets and 26 target categories. 
The list of datasets, detailed next, contains both written and spoken language, from SMS messages with informal abbreviations, through posts of movie reviews, to formal protocols and legal documents written by professionals. In addition, both clean text and noisy automatic speech recognition (ASR) output are being used. The datasets' categories, sizes and download links are provided in Appendix \ref{app:datasets}.

\paragraph{Subjectivity} \cite{ds-subjectivity:04} Subjective and objective movie reviews automatically obtained from Rotten Tomatoes and IMDb.

\paragraph{Polarity} \cite{ds-polarity:05} Positive and negative automatically derived movie reviews.

\paragraph{AG’s News} A large-scale corpus of categorized news articles. We used the description field of the version released by \citet{ds-ag-news:15}. 

\paragraph{SMS spam} \citep{ds-sms:11} SMS messages tagged for ham (legitimate) or spam.

\paragraph{\tos{}} \citep{Lippi2019CLAUDETTEAA} Terms of Service legal documents of 50 major internet sites, in which sentences were annotated for one category - whether they belong to an unfair clause. 

\paragraph{ISEAR} The International Survey on Emotion Antecedents and Reactions (ISEAR) \cite{ds-isear:15} is a collection of personal reports on emotional events, written by 3000 people from different cultural backgrounds. Each sentence in it was labeled with a single emotion (out of joy, fear, anger, sadness, disgust, shame, and guilt). 

\paragraph{\holj{}} \citep{grover2004holj} A corpus of judgments of the U.K. House of Lords: legal documents containing legal terms, references and citations from rules, decisions, and more, given as free speech. Categorized into six rhetorical roles.

\paragraph{Wiki attack} \cite{ds-wiki-attack:17} A corpus of discussion comments from English Wikipedia talk pages that were annotated for attack; personal, general aggression, or toxicity.\footnote{This data set contains offensive language. IBM abhors use of such language and any form of discrimination.} 

\paragraph{\asrd{}} Spoken debate speeches transcribed by an ASR system, as in \cite{Mirkin-etal:2018,mirkin2018listening}. We believe ASR well exemplifies a commonly used domain with scarce annotated data (especially if one considers the varieties due to different systems).

As this dataset comes with no sentence-level annotation, we created a test set by annotating 700 sentences to whether they contain an argument for a given topic. These sentences cover 20 topics with no intersection with the texts and topics from which rules were discovered.
Annotations details are given in Appendix \ref{app:asrd}, and the annotated dataset is available on the IBM Project Debater datasets webpage.\footnote{\url{http://www.research.ibm.com/haifa/dept/vst/debating_data.shtml}}
.

\paragraph{\stgur{}} \citep{stab2017parsing} Written student essays, labeled into three types of argumentative content: Major Claim, Claim, and Premise. 

\section{Evaluation} \label{sec:eval}

As described, the goal of \graspl{} is to alleviate the blank canvas problem when facing new unlabeled data, and to expedite the expert’s work. The experiments described next aim to show that the list of rules \graspl{} discovers can be useful at the hand of experts. 
We do not propose utilizing this list directly to classify sentences. 
Rather, we propose that an expert considers the list of rules and uses her expertise to gain insights and create rules for the task at hand. 
The expert can either consider a rule directly, or gain insights by looking at several sentences in the new data which a rule captures. The expert can then filter noisy rules, combine rules to create new ones, fine tune rules, and much more. Eventually, interacting with the list of rules generated by \graspl{} should help her understand the underlying categories and design rules that correspond to categories of interest.

\subsection{Simulating Expert Input}
\label{ssec:expert_simulation}

Evaluating the combination of \graspl{} with human input is a complicated task and may be noisy due to the human input. We, therefore, use a surrogate method, which assesses \graspl{} assuming a setting where the human knows or has deduced the categories based on examining the rules, and then takes a very straightforward approach, namely choosing a subset of the rules (as-is) for each category, based on their correlation to the category. 

Given the list of rules generated by \graspl{}, with no labels and no list of categories, we calculate a correlation measure  (Information Gain) between each rule and each category of the dataset on a small validation set (see below). Then, for each category we take the \sr{\textit{top k}}{\in}{\{100, 50, 25, 10\}}, rules for it, as ranked by the correlation measure.
The procedure simulates a human manually filtering rules.
We note that this simulation chooses rules independently of each other, while human experts can potentially be better in considering the dependencies between rules, combining rules and otherwise adjusting the rules. Nevertheless, this evaluation provides an estimation of what may be achieved by combining \graspl{} with human input.

Given a subset of rules, selected as above, we study whether they capture a non-trivial part of the category realizations in the data. We report the performance of using these rules to classify sentences. Our classification rule is simple - if at least \sr{x}{\in}{\{10, 5, 2, 1\}} rules match a sentence, the sentence is considered as positive.
This simulates the expert merging several rules together to increase precision. In general, a human expert is expected to outperform the simulation.

The human expert simulation is done on a validation set. For that, we randomly sampled 100 annotated sentences from each dataset.
For multi-category datasets, we sampled 300 annotations from each.
These sizes were chosen according to the number of sentences which is reasonable to expect a human expert to annotate in a limited amount of time (50--100 per category of interest).

\subsection{Experimental Setup} \label{ssec:setup}

\graspl{} has the same set of parameters as \grasp{} which can be tuned to improve performance. To keep this part simple we fix all parameters but one, which more directly affects the recall-precision trade-off (precision is deemed more important as it tilts the rules generation algorithm towards outputting more specific and informative rules). Full details are given in Appendix \ref{app:grasp-params}.

Baselines, detailed next, were tuned on the validation set. Text was vectorized as Bag of Words.

\paragraph{Prior} Choosing all instances as positive. Precision is the interesting measure to compare to here, as recall is trivially $100\%$ and meaningless.

\paragraph{SIB} SIB \citep{sib:02} is a sequential clustering algorithm that was shown to be superior to many other clustering methods \citep{sib-is-best:13}. Parameter details are found in Appendix \ref{app:full_res}. We also tried \textbf{LDA} \citep{Blei2003LatentDA}. However, it was consistently inferior to SIB and thus we only report it in Appendix \ref{app:full_res}.

\paragraph{NB} We train a Multinomial Naive-Baye classifier taking the domain corpus as the positive instances and the general English as the negative instances. Parameters are the default in the sklearn library.\footnote{\url{https://scikit-learn.org/}}

These baselines were compared to the two \graspl{} versions, according to the two options of generating the background (described in \S\ref{sec:method}):

\paragraph{\graspl{}+\geb{}} General English corpus is used as background, while the entire domain corpus (the entire dataset) is taken as foreground.

\begin{table}[!t]
\begin{small}
\begin{tabular}{@{}l|lccc@{}}
\toprule
dataset & method      & P\%  & R\%   & F$_1$\%          \\ \midrule
\multirow{6}{*}{\parbox{1cm}{SMS\\spam}}            & prior       & 13 & 100 & 23          \\
                                                                               & SIB         & 34 & 98  & 50          \\
                                                                               & NB          & 18 & 93  & 30          \\
                                                                               & \graspl{}+\geb{}    & 51 & 79  & 62          \\
                                                                               & \graspl{}+\beg{} & \textbf{93} & 73  & \textbf{82} \\ \midrule
\multirow{6}{*}{\parbox{1cm}{ToS\\ unfair\\ clause}} & prior       & 11 & 100 & 20          \\
                                                                               & SIB         & 12 & 53  & 19          \\
                                                                               & NB          & 11 & 100 & 20          \\
                                                                               & \graspl{}+\geb{}    & \textbf{25} & 42  & \textbf{32} \\
                                                                               & \graspl{}+\beg{} & 18 & 43  & 25          \\ \midrule
\multirow{6}{*}{\parbox{1cm}{Wiki\\ attack}}         & prior       & 12 & 100 & 21          \\
                                                                               & SIB         & 24 & 89  & 38          \\
                                                                               & NB          & 13 & 95  & 22          \\
                                                                               & \graspl{}+\geb{}    & 12 & 93  & 21          \\
                                                                               & \graspl{}+\beg{} & \textbf{54} & 38  & \textbf{44} \\ \midrule
\multirow{6}{*}{Subjectivity}                                                  & prior       & 52 & 100 & 68          \\
                                                                               & SIB         & \textbf{89} & 93  & \textbf{91} \\
                                                                               & NB          & 58 & 87  & 69          \\
                                                                               & \graspl{}+\geb{}    & 55 & 94  & 70          \\
                                                                               & \graspl{}+\beg{} & 79 & 79  & 79          \\ \midrule
\multirow{6}{*}{\parbox{1cm}{ISEAR \\ sadness}}      & prior       & 15 & 100 & 25          \\
                                                                               & SIB         & \textbf{59} & 22  & \textbf{41} \\
                                                                               & NB          & 15 & 83  & 26          \\
                                                                               & \graspl{}+\geb{}    & 16 & 79  & 27          \\
                                                                               & \graspl{}+\beg{} & 56 & 29  & 38 \\ \midrule
\multirow{6}{*}{\parbox{1cm}{HOLJ\\ background}}     & prior       & 41 & 100 & 58          \\
                                                                               & SIB         & 59 & 22  & 32          \\
                                                                               & NB          & 40 & 93  & 56          \\
                                                                               & \graspl{}+\geb{}     & \textbf{75} & 61  & \textbf{67} \\
                                                                               & \graspl{}+\beg{} & 57 & 76  & 65    \\ \midrule
\multirow{7}{*}{\asrd}                                                          & prior       & 36 & 100 & 53 \\
                                                                                & SIB         & 40 & 13  & 20 \\
                                                                                & NB          & 35 & 65  & 46 \\
                                                                                & BlendNet    & \textbf{52} & 32  & 40 \\
                                                                                & \graspl{}+\geb{}    & 40 & 94  & \textbf{56} \\
                                                                                & \graspl{}+\beg{} & 40 & 85  & 55 \\ \midrule
\multirow{7}{*}{\parbox{1cm}{Essays\\major claim}} & prior       & 9  & 100 & 17 \\
                                                                                & SIB         & 10 & 48  & 17 \\
                                                                                & NB          & 12 & 81  & 20 \\
                                                                                & BlendNet    & 12 & 32  & 17 \\
                                                                                & \graspl{}+\geb{}    & \textbf{32} & 65  & \textbf{42} \\
                                                                                & \graspl{}+\beg{} & 12 & 74  & 21 \\
\bottomrule     
\end{tabular}
\caption{Results of \graspl{} and the baselines on various categories, full results in Appendix \ref{app:full_res}. 
\label{tab:results}}
\end{small}
\vspace{-0.8cm}
\end{table}

\paragraph{\graspl{}+\beg{}} The foreground and background are both taken from the domain corpus. For this, we perform an in-domain split with SIB as the unsupervised clustering method.

\subsection{Results} \label{ssec:res}

As detailed in \S \ref{sec:data} we evaluate \graspl{} on 26 target categories from 10 datasets. The full results table is presented in the Appendix \ref{app:full_res}. Table \ref{tab:results} depicts representative results. The results presented for \graspl{} are the best obtained for each category after the expert simulation (See \S\ref{ssec:expert_simulation}).

On \textit{ISEAR disgust}, \textit{Polarity}, and \textit{Essays premise} no system improves over the prior baseline. On other datasets, SIB is a strong baseline, as can be seen in Table \ref{tab:results} for \textit{Subjectivity} and \textit{ISEAR sadness}. SIB also ranks first for three additional categories of \textit{ISEAR}, and all four categories of \textit{AG's news}. In all other 14 categories, at least one version of \graspl{} is ranked first.

SIB, as a bag of words method, is expected to perform well on topic classification (e.g., \textit{AG's news} dataset), but it cannot capture more subtle 
linguistic structures. \graspl{}, on the other hand, integrates signals from both the mere appearance of words in the text, as well as from the existence of more involved 
semantic structures in it. In addition, SIB by itself 
does not provide a human-readable explanation for its decisions and thus is not suitable for our scenario of assisting human experts.

As mentioned, in most cases \graspl{} outperforms the other baselines. In some cases both versions are better than the rest, e.g., \textit{SMS spam}, \textit{ToS} and \textit{\holj{} background} (see Table \ref{tab:results}).

It is more common for \graspl{}+\beg{} to outperform \graspl{}+\geb{} than the other way around (e.g., \textit{SMS spam}, \textit{Wiki attack}, and \textit{ISEAR sadness}). In some cases, \beg{} manages to achieve this superiority even though SIB, its first step, performs poorly (e.g. \textit{ISEAR fact}). But, in most such cases, SIB gains high performance and thus contributes to the superiority of \beg{} over \geb{}. 

This shows the importance of the in-domain split method. Take \textit{Wiki attack} as an example. The language and structure of its texts differ from our general English background (taken from news articles) and therefore \graspl{}+\geb{} fails to improve over the prior baseline. SIB, on the other hand, manages to outperform prior with a modest improvement in precision. This improvement is enough for \graspl{}+\beg{} to lift itself even higher. By contrasting similar texts from the same domain, it overcomes their uniqueness and more than doubles SIB precision while keeping a decent recall.

For \textit{\tos{}} dataset, \graspl{} performance is modest, probably since \textit{unfair clauses} are a small category in this data of legal documents. We hypothesise that there are other, more prominent categories in this data which are better captured by \graspl{} rules. In \S\ref{ssec:identifying_cats}, we provide an example of such rules.

For the two datasets of the computational argumentation domain (\textit{\asrd{}} and \textit{\stgur{}}), we implemented \textbf{BlendNet} \citep{BlendNet} as a competitive domain adaptation baseline. 

We train two models, one detects premises and the other claims. Train sets are proprietary datasets, each holds about 200K labeled news sentences. BlendNet predicts that an argument exists if any type of argument is detected.
The abundance of data and modern architecture make for a strong supervised baseline for comparison.\footnote{We avoid blending since it is not influential, given the amount of labeled data, as noted by the original paper.}

Considering F$_1$, we can see, in Table \ref{tab:results}, that both \graspl{} methods outrank BlendNet, the domain adapted baseline in both datasets.

To summarize, our extensive evaluation shows that in most cases \graspl{} learns useful rules for the target category in an unsupervised way. In general, while \graspl{}+\geb{} tends to prefer recall, \graspl{}+\beg{} usually favors precision. Both versions stand out in categories with low prior.
\section{Analysis} \label{sec:analysis}

After demonstrating the potential of \graspl{} in the quantitative results, we turn to a qualitative analysis. It is hard to experimentally quantify the contribution of \graspl{} rules for human experts. In \S \ref{sec:user_study} we present a user study which shows that \graspl{} model is indeed human-readable and provides explainability for its decisions.

In the next three sections, we show a recurrent ability of \graspl{} rules to capture a semantic meaning which is commonly used in a given domain, and to generalize its different formulations in it. For example, the first rule in Table \ref{tab:pattern_exmples} identifies the beginning of new parts of a speech, and can help in breaking it into meaningful sections.

\subsection{Automatically identifying categories} \label{ssec:identifying_cats}

To test our hypothesis, that \graspl{} rules capture other categories in the \tos{} dataset, rather than the low frequent target category \textit{unfair clause} (see \S \ref{ssec:res}), we conduct the following experiment.

We assigned one of the authors with the task of identifying additional categories in \tos{} (the dataset of Terms of Service legal documents), just by examining the list of rules learned for this dataset and their matching sentences.
The assignee reported learning new legal collocations and that, by merely skimming rule matches, finding their general 
context was surprisingly easy.

A prominent class of categories in the data that the assignee identified was \textit{customer side part in the agreement}. It includes categories such as \textit{what you agree to}, \textit{what you may do}, and \textit{what you must do}. Rules which identify these categories most often include terms such as \textit{you} (the customer) or \textit{we} and company names. 
For each such category, numerous rules capture different characteristics, such as matching  \textit{must}, \textit{have}, and \textit{will} or generalizing over verbs like \textit{agree}, \textit{acknowledge}, \textit{continue} and \textit{understand}. 

This analysis, although subjective, demonstrates the utility
of \graspl{} as an aiding tool when the categories underlying a new data are not known a-priori.

\subsection{\geb{} vs. \beg{}} 

Besides the differences in performance of the two methods, there are apparent qualitative differences between them. 
The \geb{} method tends to capture
words. For example, consider two examples in Table \ref{tab:pattern_exmples}; the rule for \holj{} legal domain (lines 10--12), contains the attribute \as{hyponym of written communication} which matches \textit{section} and \textit{paragraph}, and the rule for the \textit{unfair clauses} (lines 13--15) matches the word \textit{any}. In first sight, the last rule is deemed trivial. However, the word ``any'' did stand out and appeared in many rules. When inspecting a couple of sentences that match this rule, 
it is apparent that they often convey strong 
statements with an inclusive phrasing (e.g., \textit{we will not be held liable for \textbf{any} disruption of service}).

On the other hand, the \beg{} method may capture specific words as well, but mostly it generalizes (e.g., \as{hyponyms of rank}) or, more often, relies on abstract notions, 
expressed through syntax, WordNet and the sentiment lexicon.  

These findings are in line with the hypothesis that the dissimilarity between a domain foreground and a general English background may lead to over-reliance on jargon words. Thus, emphasizing the need for the in-domain split method. However, rules containing common words are still effective for capturing indications similar to those other unsupervised methods, such as NB, capture. 

Inspecting the failures of \geb{} 
reveals another issue with this method.
In the \textit{fact} category, for example, sentences are short laconic statements. This is unique in comparison to the rest of \holj{} corpus, but not in comparison to general English. So, their dissimilarity to the rest of the corpus is found only in \beg{}. This is also the case for another fail in \textit{Framing}. It might be the case that adding attributes (e.g., sentence length or a measure of structural complexity) or extracting a larger set of rules would alleviate the problem.

\subsection{A knowledgeable in-domain split reveals known findings in the literature} \label{ssec:split_by_indicator}

When describing the in-domain split in \S \ref{sec:method} we mention a knowledgeable in-domain split for the computational argumentation domain, i.e., taking the first halves of sentences as the foreground and the rest as the background. We next show that rules learned with this heuristic capture known findings in the computational argumentation domain. 

In \stgur{} annotation guidelines, \citet{stab2017parsing} provide two lists of indicators for claims and premises to facilitate the annotation task of identifying these categories.  

We found out that \graspl{}, applying the above mentioned knowledgeable in-domain split, produces rules which capture these indicators and generalize them. By examining rules matches in the corpus, one can easily obtain additional specifications of these indicators. For example, lines 7--8 in Table \ref{tab:pattern_exmples} show that the third rule captures two premise indicators stated in the guidelines, \textit{for example} and \textit{for instance}. Line 9 shows that it also captures indicators not listed there, such as \textit{as a matter of fact} and \textit{in fact}.

\subsection{User Study}\label{sec:user_study}
One of the advantages of \graspl{} is that it is an explainable model, making predictions based on rich and interpretable rules. These can be used to justify predictions, sometimes termed a local explanation \cite{lertvittayakumjorn-toni-2019-human} and also to understand the way the model works as a whole (termed global explanation), potentially enabling experts to build better classifiers. 

We performed a user study aimed at studying whether \graspl{} is viewed as interpretable by human users. We focus on the question of local explanation, namely when considering a specific instance, does examining the rules matched by \graspl{} help the user understand \textbf{why} the model made the prediction (as opposed to assessing whether it is a model that will produce good predictions). The study was conducted on the SMS spam dataset since it is a familiar task for users. 

Following \citet{models-comparison:19}, we designed a comparative study in which an example is presented with two explanations (A and B), and the user is asked to choose which one better explains how the system made its prediction. We chose NB as the comparative model, because like \graspl{}, it is an unsupervised model, and can output an explanation in the form of indicative keywords. 
To eliminate precision differences between the methods, we randomly sample examples which both methods correctly recognized as spam messages and presented 20 examples.\footnote{Preliminary experiments showed that to get a view of user preference a limited number of examples suffices.}
Given a text sequence identified as spam by both models, NB's explanation is the list of words that were found to be strongly related to spam. Analogously, \graspl{} explanation is a list of rules that were matched in the text sequence (see screenshot in Appendix \ref{app:user_study}). The order in which model explanations appear in each example (i.e., which one is A) is random. We used $7$ annotators for this study.
The full guidelines and users' aggregated annotations are found in Appendix \ref{app:user_study}. 

We ignored one outlier that was too positive towards \graspl{}. Overall, in $53\%$ of the times, users preferred \graspl{} explanations ($41\%$ of those were with a strong preference). In $29\%$ they abstained and in only $18\%$ of the times NB explanation was considered better than that of \graspl{}. 

In summary, although this is an anecdotal experiment, it shows that the fact that \graspl{} model is rich and interpretable is useful for interaction with humans, and allows them to better understand a model's prediction, when compared to words only. We leave for future work the interesting topic of how one can use \graspl{} as a surrogate model over black-box models, as well as how an expert may utilize the rules offered by \graspl{} to efficiently build rule-based models.
\section{Related Work} \label{sec:bg}

Our work provides a method to explore new data. In statistics, the field of analyzing new datasets is called Exploratory Data Analysis \cite{yu1977exploratory, fekete2016progressive}. In NLP, such work is less common and characteristics of each dataset, task or domain are extracted independently \cite{choshen2018inherent, koptient2019simplification}. This has the benefit of gaining a deep understanding of each task. For instance, the work on translation divergences \citep{dorr-1994-machine, nidhi2018english} that aims to better explain translation to support system development later on. 

Research about patterns and expert crafted rules was popular in the past \citep{Hearst1992AutomaticAO,Kukich1992TechniquesFA, Ravichandran2002LearningST} and is still found useful nowadays; for enhancing embeddings \citep{Schwartz2017patterns},
filtering noise in crawled data \citep{grundkiewicz2014wiked,koehn2019findings}, as a component within large pipelines \citep{ein2019corpus} or by itself in text-rich domains \cite{Padillo2019EvaluatingAC}. Using domain expertise to categorize and understand a new domain is often the first practical step to apply in other fields too, which may devise rules for that purpose  \cite{Brandes2018PostConsumptionS, pschy2019verbal, 
 Li2019MDBADM, Nguyen2010SymbolicRC}.

With the increasing use of AI, a new field is emerging -- Explainable AI (XAI). It is concerned with how to understand models' inner workings. LIME \citep{ribeiro:kdd16} attempts to explain predictions by perturbing the input and understanding how the predictions change. Other works use attention as a mechanism to interpret a model's prediction (see e.g., \citealp{ghaeini-etal-2018-interpreting}, who propose to interpret the intermediate layers of DNN models by visualizing the saliency of attention and LSTM gating signals). A survey of the XAI field for NLP does not exist but see \cite{gilpin2018explaining,alej2019explainable} for surveys of the XAI field in general. We show in this paper that \graspl{} is interpretable by human users and is thus interesting for the XAI community.
\section{Conclusions} \label{sec:conclusions}
We present \graspl{}, an unsupervised, explainable method, which does not require substantial computing resources, and can expedite the work of human experts when approaching new datasets. We describe two methods for obtaining the background and foreground corpora which \graspl{} relies on, and compare them. We note that our method is not limited to any specific language. All \graspl{} needs is a few basic text processing tools. 

Examining numerous datasets, we demonstrate that with no labeled data, nor any information about the categories underlying these datasets, \graspl{} is able to identify indicative rules for a wide variety of categories of interest. Our analysis shows that these rules often capture a common semantic meaning which can be realized in many different ways in the data. Finally, a user study further shows that these expressive rules provide valuable explanations for classification decisions.

Finally, the fact that \graspl{} was found useful for most of the 26 categories on which it was evaluated (despite their difference) increases our belief that it can be very practical for your next dataset.


\FloatBarrier
\bibliography{bibi}
\bibliographystyle{acl_natbib}

\FloatBarrier
\appendix
\section{Datasets} \label{app:datasets}
\begin{description}

    \item [AG’s News:] \url{http://groups.di.unipi.it/~gulli/AG_corpus_of_news_articles.html}.\\
    We used the version from: \url{https://pathmind.com/wiki/open-datasets} (look for the link \textit{Text Classification Datasets}).
    
    \item [\asrd:] \url{https://www.research.ibm.com/haifa/dept/vst/debating_data.shtml} (look for the \textit{Debate Speech Analysis} section).

    \item [\stgur:] 
    \url{https://www.informatik.tu-darmstadt.de/ukp/research_6/data/index.en.jsp}
    
    \item [\holj:]
    \url{https://www.inf.ed.ac.uk/research/isdd/admin/package?download=84}
    
    \item [ISEAR:]  \url{https://www.unige.ch/cisa/research/materials-and-online-research/research-material/}.
    
    \item [Polarity:] \url{http://www.cs.cornell.edu/people/pabo/movie-review-data/}.
    
    \item [SMS spam:] \url{http://www.dt.fee.unicamp.br/~tiago/smsspamcollection/}
    
    \item [Subjectivity:] \url{http://www.cs.cornell.edu/people/pabo/movie-review-data/}. 
    
    \item [\tos:]
    \url{http://claudette.eui.eu/ToS.zip}

    \item [Wiki attack:] \url{https://figshare.com/articles/Wikipedia_Talk_Labels_Personal_Attacks/4054689}.
    
\end{description}

\begin{table*}[!t]
\centering
\resizebox{\textwidth}{!}{%
\begin{tabular}{@{}llcccccc@{}}
\toprule
Dataset                    & Category      & Train Size              & Train Prior & Validation Size      & Validation Prior & Test Size              & Test Prior \\ \midrule
\multirow{4}{*}{AG's news} & world         & \multirow{4}{*}{10,000} & 0.24        & \multirow{4}{*}{300} & 0.25             & \multirow{4}{*}{3,000} & 0.25       \\
                           & sports        &                         & 0.26        &                      & 0.23             &                        & 0.25       \\
                           & business      &                         & 0.25        &                      & 0.25             &                        & 0.26       \\
                           & sci/tech      &                         & 0.25        &                      & 0.28             &                        & 0.25       \\ \midrule
\multirow{3}{*}{Essays}    & claim         & \multirow{3}{*}{5,303}  & 0.10        & \multirow{3}{*}{300} & 0.10             & \multirow{3}{*}{1,344} & 0.09       \\
                           & major claim   &                         & 0.51        &                      & 0.53             &                        & 0.56       \\
                           & premise       &                         & 0.20        &                      & 0.19             &                        & 0.19       \\ \midrule
\multirow{7}{*}{HOLJ}      & background    & \multirow{7}{*}{844}    & 0.07        & \multirow{7}{*}{300} & 0.06             & \multirow{7}{*}{544}   & 0.07       \\
                           & disposal      &                         & 0.18        &                      & 0.19             &                        & 0.18       \\
                           & fact          &                         & 0.18        &                      & 0.19             &                        & 0.18       \\
                           & framing       &                         & 0.06        &                      & 0.06             &                        & 0.07       \\
                           & proceedings   &                         & 0.40        &                      & 0.38             &                        & 0.41       \\
                           & textual       &                         & 0.10        &                      & 0.12             &                        & 0.10       \\
                            \midrule
\multirow{7}{*}{ISEAR}     & anger         & \multirow{7}{*}{5,366}  & 0.14        & \multirow{7}{*}{300} & 0.13             & \multirow{7}{*}{1,534} & 0.14       \\
                           & disgust       &                         & 0.14        &                      & 0.11             &                        & 0.16       \\
                           & fear          &                         & 0.14        &                      & 0.20             &                        & 0.15       \\
                           & guilt         &                         & 0.14        &                      & 0.13             &                        & 0.14       \\
                           & joy           &                         & 0.15        &                      & 0.13             &                        & 0.13       \\
                           & sadness       &                         & 0.14        &                      & 0.13             &                        & 0.15       \\
                           & shame         &                         & 0.14        &                      & 0.16             &                        & 0.13       \\ \midrule
ASRD                       & argument      & 10,378                  & 0.37        & 100                  & 0.37             & 600                    & 0.37       \\ 
Polarity                   & positive      & 7,463                   & 0.50        & 100                  & 0.51             & 2,133                  & 0.50       \\ 
SMS spam                   & spam          & 3,900                   & 0.13        & 100                  & 0.12             & 1,115                  & 0.13       \\
Subjectivity               & subjective    & 7,000                   & 0.50        & 100                  & 0.54             & 2,000                  & 0.52       \\
ToS                        & unfair clause & 9,414                   & 0.11        & 100                  & 0.09             & 9,314                  & 0.11       \\
Wiki                       & attack        & 10,000                  & 0.11        & 100                  & 0.09             & 3,000                  & 0.12       \\ \bottomrule

\end{tabular}%
}
\caption{Statistics for the used datasets. Prior refers to the percentage of the target category examples in the data.\label{tab:dataset_stats}}
\end{table*}

We present in Table \ref{tab:dataset_stats} the number of examples in each dataset part (i.e., train, dev, and test) for each target category, together with the percentage of examples from the target category (the prior).

\section{Annotating \asrd{}} \label{app:asrd}
Each sentence of \asrd{} was annotated by three expert annotators who are fluent English-speakers with long experience in argumentation tasks. Each sentence was presented within a context from the speech and its topic. Annotators were asked whether it contains an argument for the given topic. Their majority vote was taken as the label. 

The average pairwise Cohen's kappa \cite{Cohen1960} between annotators is 0.35 (a typical value in computational argumentation tasks, e.g., \citealp{aharoni14,rinott15}). The prior for positive in the test set is 0.37.

\subsection{\asrd{} Test Set Annotation Guidelines}
These are the guidelines provided to the annotators:

In the following task you are given a part of a transcription of a spoken speech delivered over a controversial topic. Note, the transcription is often done automatically, hence may contain errors (such as wrong transcription of words, bad split of the speech into sentences). Try to figure out what the speaker really said and base your decisions on that.

A sentence is given with its context in the speech. For this sentence you should determine whether it contains an argument for the given topic.

An argument is a piece of text which directly supports or contests the given topic.
Note: having a clear stance towards the topic (either pro or against) is a critical prerequisite for a piece of text to be an argument.

\section{\grasp{} Parameters} \label{app:grasp-params}
To extract \grasp{} attributes we used \href{https://opennlp.apache.org/}{OpneNLP} POS tagger, \href{https://nlp.stanford.edu/software/CRF-NER.shtml}{Stanford NER}, WordNet hypernyms and \href{https://wordnet.princeton.edu/documentation/lexnames5wn}{super-classes}, and \newcite{sentimentLexicon:04}  sentiment lexicon.

We report the parameters used for the \grasp{} algorithm (notations follow the ones defined in \grasp{} paper). This configuration is by no means the optimal one:

\begin{itemize}
\item Size of the alphabet \sr{k_1}{=}{1000}
\item Number of rules to learn \sr{k_2}{=}{100}
\item Max rule length (in attributes) \sr{maxLen}{=}{5}
\item Rules correlation threshold \sr{t_2}{=}{0.5}
\item Rule match window size \sr{w}{=}{5}
\item Min freq of attribute in data \sr{t_1}{=}{0.005}
\end{itemize}
These parameters are kept fix during all experiments. Another parameter of \grasp{} is the scoring function used to rank attributes and rules during learning. We chose F$_\beta$ (as opposed to the original Info Gain) which allows us to tune between recall and precision. As mentioned in the paper, we prefer giving a higher weight for the foreground. Therefore, we try $\beta \in \{0.5, 0.1, 0.05\}$ which makes this scoring function asymmetric with a preference for precision. The different values were chosen without any deep thought to cover three precision orientation levels - small, medium, and large. 

\graspl{} does not demand special hardware and can be run on a normal laptop in a reasonable amount of time.

\section{Full results and configuration} \label{app:full_res}
In this section we report more baselines we ran and their tuning and the full results table, Table \ref{tab:full_res}.

\textbf{SIB} - 
We used 10 restarts, each with a random partition of equally populated clusters and then apply up to 15 optimization iterations. Early stop happens when the number of elements that switched clusters was less than 2\% of the total elements. We assume uniform prior on the data, which means that all texts have equal probability.

\textbf{LDA} - Latent Dirichlet Allocation \citet{Blei2003LatentDA} is a very common unsupervised method for topic classification. We utilize the sklearn library.\footnote{\url{https://scikit-learn.org/stable/}} 
We set the number of clusters to be the number of categories per dataset (a piece of information which is not provided to \graspl{}). This choice was consistently better than setting a larger number of clusters. We also performed a grid search over the validation set of hyper parameters, but the best performance was obtained by choosing the default parameters in the sklearn library. Despite trying hyperparameter tuning on the test set LDA results were low and we hence resorted to include only the stronger unsupervised baselines in the paper.

\subsection{Supervised experiment}

In addition to the obvious baselines we add the context of supervised methods and show results of BERT \citep{Devlin2019BERTPO} as probably the strongest supervised classification system. We note that since BERT's model is not interpretable it is not suitable for our scenario, in which explainability is needed to assist human experts, it is also not an unsupervised method despite its high performance on small amounts of data. It is important to note that despite the use of development sets to simulate a human, the unsupervised methods in the paper are indeed unsupervised and supervised methods are expected to have higher performance whenever possible (e.g. \grasp{} would outperform \graspl{}). 
We report the performance of supervised methods here, as to not withhold the information gathered in the experiments.

\textbf{BERT} - we fine-tune BERT on the validation set, choosing the best model after 5 epochs. With small training sizes, BERT performance fluctuates even more than usually reported \cite{Dodge2020FineTuningPL}, therefore we report average of 3 runs. Also note that while for some datasets  there are seeds for which BERT classifies everything as the common label, for \tos{} we could not achieve a run with meaningful classification, despite 9 trials.

Another supervised method we compared to is \textbf{NB-on-dev} in which we train Multinomial Naive-Bayes as a supervised classifier over the validation. Parameters were the default in the sklearn library.

The full results are given in Table \ref{tab:full_res}.
It is not surprising that on most dataset supervised methods perform quite well. Although, this is more the case with BERT than the case with NB-on-dev which often underperform \graspl{}. Some may even say that it is surprising that unsupervised methods are anywhere close to the supervised ones, this is probably explained by the paucity of data for training.

\section{Human in the Loop Parameters} \label{app:human_in_loop}

In the result section we report the best performance per category and foreground / background method. These results were obtained after simulating the human expert in the loop. Beyond choosing top rules, \textit{topK}, by the correlation measure, we also maximized over two parameters that are considered to be tuned by the expert: (i) \textit{min rules matches} - how many rules should be matched in a candidate sentence for it to be considered positive for the category, and (ii) $\beta$ value for F$_\beta$ which reflects expert's preference in the recall--precision trade off.

The parameters with which the best performance was obtained for each category and background method are found in Table \ref{tab:full_res}.

\section{User Study} \label{app:user_study}
In this sections we provide the guidelines for the user study. Table \ref{tab:hum_study} depicts the all judgments of the annotators.

\begin{figure*}
    \centering
    \includegraphics[width=\textwidth]{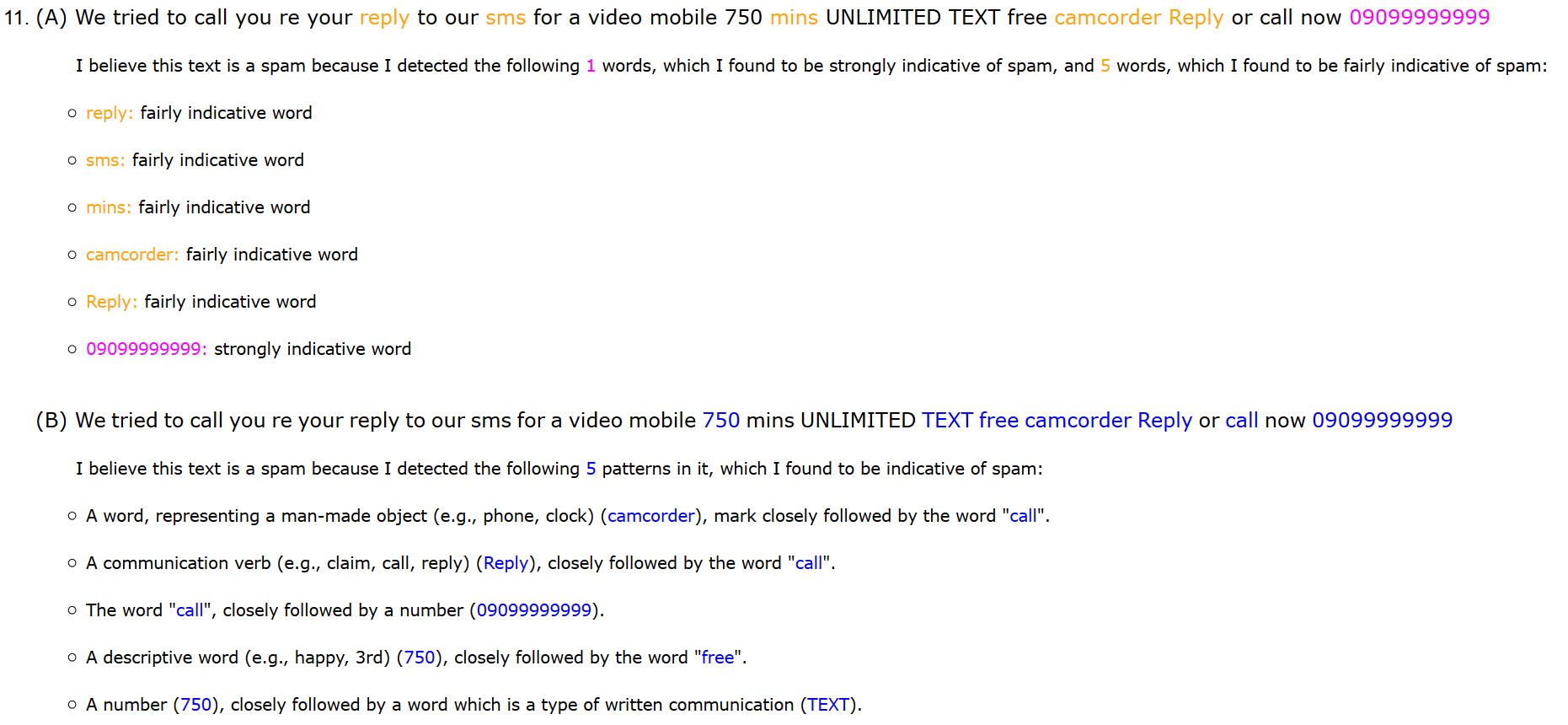}
        \caption{A screenshot of one of the sentences presented in the user study. In this sample grasp was randomly selected to appear second (B).}
    \label{fig:hum_study}
\end{figure*}

Fig. \ref{fig:hum_study} is a screenshot of a single annotation example which we manually anonimyze, as the spam dataset contains real numbers, names and addresses. 
Naive Bayes \textit{strongly indicative} and \textit{fairly indicative} words were chosen by threshold of the per word probability. The threshold were manually fitted to provide enough representative words in each sentence but avoiding having too many as too look uninformative, due to coloring all of the sentence. The chosen thresholds were more than 0.85 for strongly indicative words, and more than 0.7 for fairly indicative words.

\subsection{Guidelines}
These are the guidelines provided to the annotators:\\
\\
In this task, you are presented with spam SMS messages that were correctly identified as such by an automatic system. For each message, the system provides two explanations (A and B) for its decision. You should annotate when one explanation is preferred by you over the other \textbf{in explaining how the system works}.

Note that we are \textbf{not} interested in which explanation you think will produce better predictions of spam on new texts. Our goal is different, we want the system to produce an explanation that clarifies \textbf{why} it classifies a text as spam.

For example, a completely “black box” system giving an explanation like “I learned a model that produced 100\% accuracy on many texts, so I am confident about my predictions” should score low, because although you may believe the system produces good predictions, you cannot understand how it “knows” what is spam.

You should choose between: Definitely A, Rather A, Difficult to say, Rather B, or Definitely B.

\begin{table*}[bth]
\centering
%

\caption{Judgments per annotator of the explainabillity of \graspl{} vs. NB \label{tab:hum_study}}
\end{table*}

\onecolumn

\begin{longtable}{@{}l|lccccccc@{}}
\toprule
dataset-category                         & method           & P\%  & R\%   & F$_1\%$ & \begin{tabular}[c]{@{}c@{}}with \\ surface \\ form?\end{tabular} & F$_\beta$ & \begin{tabular}[c]{@{}c@{}}top K \\ patterns \\ in use\end{tabular} & \begin{tabular}[c]{@{}c@{}}min \\ matches \\ for positive\end{tabular} \\* \midrule
\endfirsthead
\multicolumn{9}{c}%
{{\bfseries Table \thetable\ continued from previous page}} \\
\toprule
dataset-category                         & method           & P\%  & R\%   & F$_1$\% & \begin{tabular}[c]{@{}c@{}}with \\ surface \\ form?\end{tabular} & F$_\beta$ & \begin{tabular}[c]{@{}c@{}}top K \\ patterns \\ in use\end{tabular} & \begin{tabular}[c]{@{}c@{}}min \\ matches \\ for positive\end{tabular} \\* \midrule
\endhead
\bottomrule
\endfoot
\endlastfoot
\multirow{7}{*}{AG's news business}      & LDA              & 27 & 35  & 31    &                                                                  &      &                                                                     &                                                                        \\
                                         & NB               & 25 & 76  & 37    &                                                                  &      &                                                                     &                                                                        \\
                                         & \graspl{}+\geb{}  & 26 & 96  & 41    & N                                                                & 0.5  & 100                                                                 & 5                                                                      \\
                                         & prior            & 26 & 100 & 41    &                                                                  &      &                                                                     &                                                                        \\
                                         & \graspl{}+\beg{} & 65 & 71  & 68    & Y                                                                & 0.1  & 100                                                                 & 1                                                                      \\
                                         & NB on dev        & 67 & 74  & 70    &                                                                  &      &                                                                     &                                                                        \\
\end{longtable}
\twocolumn

\end{document}